\documentclass{article}

\usepackage[nonatbib,preprint]{neurips_2019}
\usepackage[utf8]{inputenc} 
\usepackage[T1]{fontenc}    
\usepackage{hyperref}       
\usepackage{url}            
\usepackage{booktabs}       
\usepackage{amsfonts}       
\usepackage{nicefrac}       
\usepackage{microtype}      
\usepackage{graphicx}
\usepackage[noend]{algpseudocode}
\usepackage{algorithmicx,algorithm}
\usepackage{xcolor}

\title{SCAN: A Scalable Neural Networks Framework Towards Compact and Efficient Models}

\author{
  Linfeng Zhang\\
  Tsinghua University\\
  \texttt{zhanglinfeng1997@outlook.com} \\
  \And
  Zhanhong Tan\\
  Tsinghua University\\
  \texttt{tanzhh515@foxmail.com} \\
  \And
  Jiebo Song\\
  IIISCT\\
  \texttt{songjb@iiisct.com}\\
  \And
  Jingwei Chen\\
  Hisilicon\\
  \texttt{jean.chenjingwei@hisilicon.com}
  \And
  Chenglong Bao\\
  Tsinghua University\\
  \texttt{clbao@mail.tsinghua.edu.cn}
  \And
  Kaisheng Ma\\
  Tsinghua University\\
  \texttt{kaisheng@mail.tsinghua.edu.cn}
}
\begin{document}

\maketitle

\begin{abstract}
  Remarkable achievements have been attained by deep neural networks in various applications. However, the increasing depth and width of such models also lead to explosive growth in both storage and computation, which has restricted the deployment of deep neural networks on resource-limited edge devices. To address this problem, we propose the so-called SCAN framework for networks training and inference, which is orthogonal and complementary to existing acceleration and compression methods. The proposed SCAN firstly divides neural networks into multiple sections according to their depth and constructs shallow classifiers upon the intermediate features of different sections. Moreover, attention modules and knowledge distillation are utilized to enhance the accuracy of shallow classifiers. Based on this architecture, we further propose a threshold controlled scalable inference mechanism to approach human-like sample-specific inference. Experimental results show that SCAN can be easily equipped on various neural networks without any adjustment on hyper-parameters or neural networks architectures, yielding significant performance gain on CIFAR100 and ImageNet. Codes will be released on github soon.

\end{abstract}

\section{Introduction~\label{section_introduction}}
Recently deep learning has evolved to become one of the dominant techniques in areas like natural language processing~\cite{bert,attentionnlp} and computer vision~\cite{liu2016ssd,pyramidfeature}.
To achieve higher accuracy, over-parameterized models~\cite{vgg,ResNeXt} have been proposed at the expense of explosive growth in storage and computation, which is not available for certain application scenes such as self-driving cars and mobile phones. Various techniques have been utilized to address this problem, including pruning~\cite{deepcompression,pruning}, quantization~\cite{binary,xnor}, lightweight neural networks~\cite{mobilenets} design and knowledge distillation~\cite{distill_hinton,fitnets}.

Another rising star in this domain named scalable neural networks has attracted increasing attention due to its effectiveness and flexibility. The scalability of neural networks refers to its ability to adjust the trade-offs between response time and accuracy on the fly.
As a result, scalable neural networks can always accomplish inference in budgeted and limited time, which is important for real-world applications.
Researchers have explored scalability through the lens of depth (layers) and width (channels).
Built upon DenseNet~\cite{huang2017densely}, MSDNet ~\cite{multiscaledensenet} directly trains multiple classifiers from features at different levels according to their depths. Yu ~\emph{et al.} proposes switchable batch normalization which enables neural networks to work with arbitrary channels~\cite{slimmable}. However, most existing scalable neural networks still suffer from two drawbacks. Firstly, in MSDNet, multiple classifiers which share the same backbone neural network interfere with each other, leading to accuracy loss compared with training them individually. Secondly, in slimmable neural networks, computation of narrow classifiers can't be reused by wide classifiers, which means the inference of wide classifiers has to predict from scratch, increasing the inference time.

In this paper, we propose SCAN, a scalable neural network framework to overcome aforementioned difficulties.
By dividing neural networks according to its own depth straightforwardly, the computation of each classifier can be shared.
Through knowledge distillation and attention mechanism, multiple classifiers in the same backbone networks can benefit from each other instead of creating negative interaction.
Substantial experimental results show that SCAN is a generic and effective neural networks framework which can be easily equipped with various neural networks without any adjustment in architectures or hyper-parameters.

The proposed SCAN framework is inspired by human vision systems.
When a human being is asked to identify some images, most of easy images can be recognized instantly. Only fuzzy or easy-to-mix challenging images require further consideration.
In SCAN framework, images which are easy to be classified are predicted by shallow (shallow classifier) classifiers, consuming extremely little computation.
Deep classifiers only involve in the prediction of challenging samples.
Compared to traditional neural networks in which all the samples are treated with equal efforts, SCAN can obtain a high ratio acceleration through human-like sample-specific dynamic inference.

The contributions of this paper are summarized as follows:
\begin{itemize}
  \item To the best of our knowledge, this is the first work combining model compression and acceleration with attention mechanism, which provides a novel choice for lightweight models design. Compared to existing lightweight design, the
  proposed mechanism is more hardware friendly with features of input-specific adaptive networks and reuse of the backbone.
  To verify its effectiveness and generalization, SCAN is evaluated on various neural networks and datasets without adjustment in hyper-parameters and networks architectures.
\item Through this method, a two-fold improvement can be achieved on either accuracy or acceleration. Firstly, significant accuracy gain can be observed, especially on the shallow classifiers. Secondly, a high ratio of acceleration can be obtained via scalable inference mechanism.

\end{itemize}

\vspace{-0.3cm}
\section{Related work\label{section_related work}}
\textbf{Adaptive computation graph:}
Adaptive computation graph is proposed to attain flexible and dynamic neural networks acceleration~\cite{multiscaledensenet,droplayer,skipnet,blockdrop}. Compared with constant and static computation graph, it can meet various demands from diverse application scenes and inputs. SkipNet ~\cite{skipnet} targets to skip redundant layers in over-parameterized models like ResNet. A reinforcement learning based on auxiliary gating module is proposed to decide whether to skip or not. BlockDrop~\cite{blockdrop} refines this method by producing a sample-specific dropping strategy, further enhancing acceleration.

However, serious complication hides behind. Complex construction of skipping or dropping path severely prevents hardware from better coordination, which makes aforementioned algorithm counterproductive. To address this problem, the proposed SCAN simply divides the neural networks into 3 or 4 sections, which can be constructed as static graphs individually. The scalability and flexibility only exist among them instead of inside them, which is not only harmonious with hardware but also gains a higher acceleration ratio.

\textbf{Attention mechanism:}
Attention mechanism of neural networks has been extensively utilized in various fields of deep learning, yielding state-of-the-art results. It facilitates neural networks to focus on valuable information of inputs, aiming to avoid interference from redundant messages.
Firstly, proposed in machine translation, attention mechanism aims to align the words in the source language and target language~\cite{attentionnlp}. Then, it has been applied in other application of natural language processing~\cite{attentionisallyouneed, wang2017dynamic, bert}, evolving to an indispensable module in neural networks architectures. Motivated by its success in NLP, attention mechanism has also been employed in computer vision tasks.
Conspicuous performance gain has also been observed in images recognition~\cite{attentionresidual}, images caption~\cite{xuk2015show}, and fine-grained classification~\cite{lookcloser}.

In this paper, a simplified squeezing-expansion attention module has been proposed to facilitate the training of shallow classifiers, improving the accuracy of shallow classifiers in various networks.

\textbf{Model compression:}
The phenomenon that over-parameterized models can’t be deployed on edge devices for their excessive requirements of storage and computation has stimulated research of models compression and acceleration. The typical methods include pruning, quantization, compact models design and knowledge distillation. Pruning~\cite{deepcompression, pruning} is to cut off the redundant connections or channels in pre-trained neural networks. Quantization~\cite{xnor, binary} targets at replacing the 32 bits float numbers with fewer bits. Knowledge distillation~\cite{distill_hinton, distillationorigin, deepmutuallearning, attentiondistillation} aims to transfer the knowledge of over-parameterized models to a small model in order to approach higher accuracy with fewer parameters and computation. In addition, some researchers try to design compact models~\cite{mobilenets,squeezenet} which has fewer parameters yet still a high accuracy.

Aforementioned work has thoroughly exploited the redundancy in weights and neural networks architectures yet ignored the drawback that samples with diverse difficulty are treated equally.
The proposed SCAN framework is orthogonal and complementary to aforementioned work, targeting at exploring more acceleration possibility through unbalanced sample-specific disposal.

\begin{figure}[htbp]
    \vspace{-0.25cm}
    \centering
    \includegraphics[width=13.447cm,height=5.985cm]{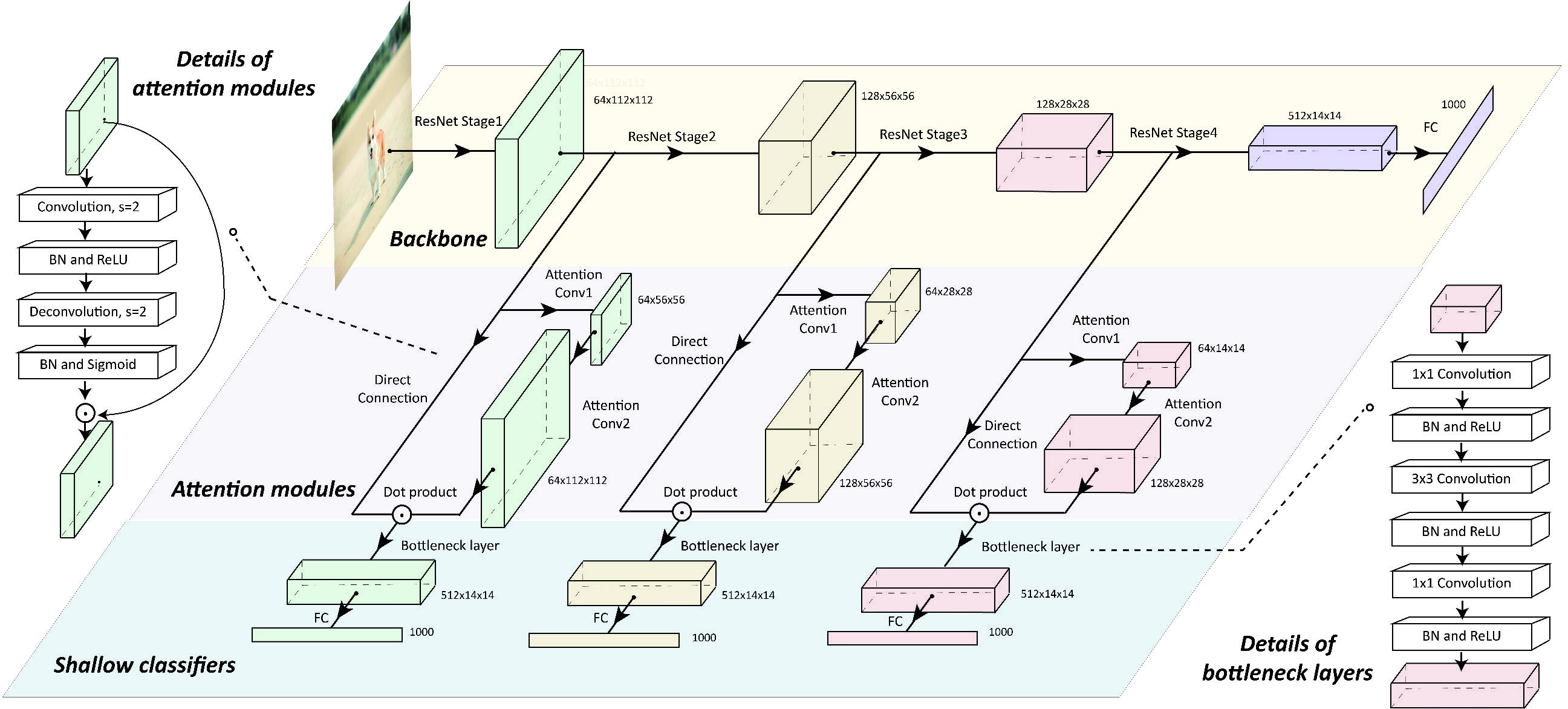}
    \caption{The architecture of ResNet18 equipped with SCAN. (i) The whole neural networks can be divided into three sections: backbone, attention modules and shallow classifiers. (ii) The backbone section is just identical to the origin model. (iii) Additional attention modules are attached after the intermediate features of backbone.(iv) Features refined by attention modules will be feed into the shallow classifiers, which consist of a bottleneck layer and a fully connected layer.}
    \label{fig:figure1}
\end{figure}

\section{SCAN framework\label{section_scannet}}
In this section, we introduce the proposed SCAN architecture as 3 parts, as shown in Figure \ref{fig:figure1}.
Firstly, to obtain human-like scalable inference, classifiers with varying response time are indispensable. According to self distillation~\cite{selfdistillation}, a bottleneck layer and a fully connected layer are organized as shallow classifiers. In addition, knowledge distillation is utilized to facilitate the training of shallow classifiers, which will be further introduced in Section~\ref{sd}.

Secondly, although shallow classifiers permit instant prediction, it also leads to dramatic decline on accuracy. To address this problem, a simplified attention module is proposed to compensate for the accuracy of shallow classifiers, which will be further introduced in Section~\ref{subsection_att}.

We further propose a threshold-based strategy to manage all the shallow classifiers to corporate together. Moreover, a genetic algorithm is designed to search for proper thresholds, which will be brought forth in Section ~\ref{scalable inference mechanism}.


\subsection{Self distillation\label{sd}}
Self distillation provides an effective method to construct and train shallow classifiers which share the same backbone neural network.
It firstly divides neural networks into several sections depending on their depth. Then a bottleneck layer and fully connected layer are attached after the intermediate features as shallow classifiers, which are regarded as the student models in knowledge distillation.
In the training period, the knowledge of the deepest classifier is distilled into each shallow classifier, whose function loss can be written as
\begin{equation}
loss = \sum_{i=1}^C loss_i
\\= \sum_{i=1}^C\Big((1-\alpha)\cdot Cross Entropy (q^i,y)
\\+ \alpha \cdot KL(q^i,q^C) + \lambda \cdot ||F_i - F_C||^2_2 \Big)
\label{equation1}
\end{equation}
where $C$ denotes the number of classifiers. $q^i$ and $q^C$ represent the outputs of softmax in the $i_{th}$ classifier and the deepest classifier respectively. $y$ represents corresponding labels. $F_i$ and $F_C$ signify the feature maps in the $i_{th}$ classifier and the deepest classifier respectively. $CrossEntropy$, $KL$ denote the well-known cross entropy loss and Kullback–Leibler divergence respectively. Experiments results show that significant performance gain can be observed on not only shallow classifier but also the deepest classifiers. Motivated by its impressive achievement, self distillation is utilized to construct and train the shallow classifiers in SCAN.

\begin{figure}
    \vspace{-0.2cm}
    \centering
    \includegraphics[width=11.2cm,height=4cm]{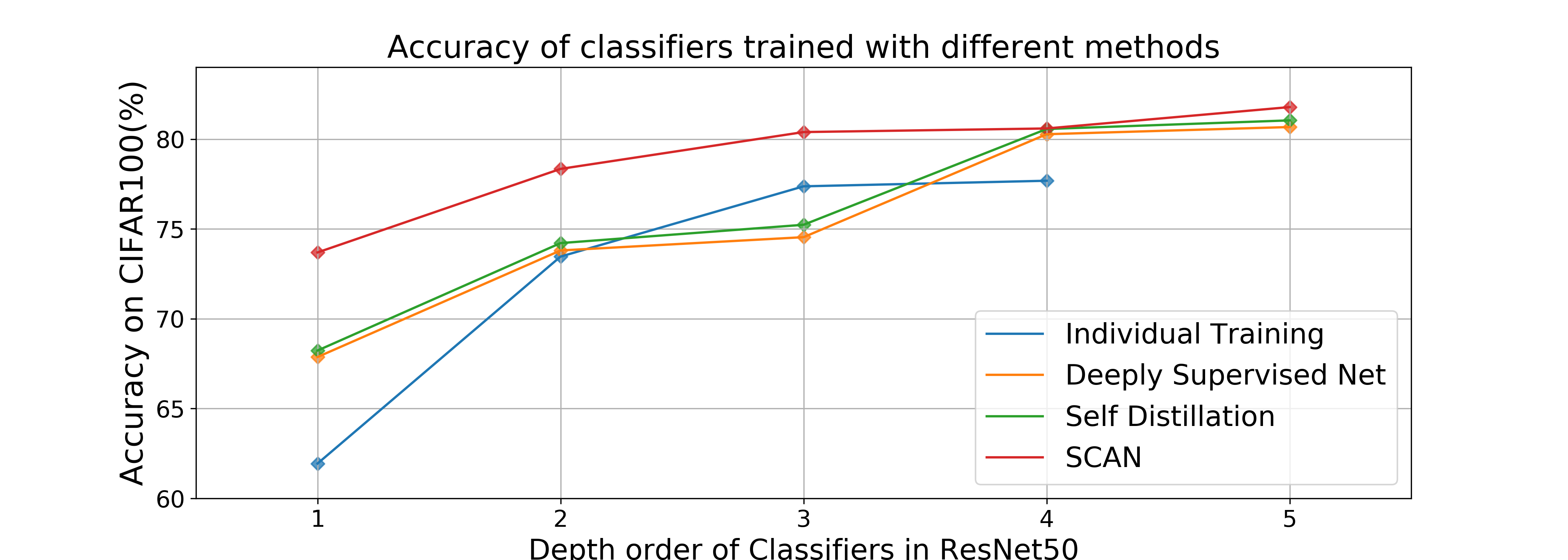}
    \caption{The accuracy of classifiers trained with different methods.}
    \label{fig:accuracy trend}
    \vspace{-0.3cm}
\end{figure}

\subsection{Attention modules\label{subsection_att}}
Figure \ref{fig:accuracy trend} provides accuracy comparison of four methods training shallow classifiers in ResNet50 on CIFAR100. The X axis is the depth of classifiers, where x=5 indicates the ensemble of all the classifiers, and Y axis denotes accuracy. It's observed that evident accuracy decay can be observed as the depth of classifiers decreases. For example, 13\% and 8\% drop on accuracy exists on the shallowest and second shallowest classifier in self distillation.
Moreover, as depicted in Figure \ref{fig:accuracy trend}, the $3_{th}$ classifier of self distillation and DSN~\cite{deeplysupervisednet} is lower than individual training, which may be caused by the negative interaction among classifiers in one backbone neural network ~\cite{multiscaledensenet}. Features desired by different classifiers are mixed up in the sharing backbone neural network. It's unattainable for each classifier to detach its own features automatically.
\begin{algorithm}
\caption{Scalable Inference\label{algorithm1}}
\hspace*{0.02in} {\bf Input:}  \hspace*{0.02in}
Samples $X$, Thresholds $\Sigma = \{\sigma_{i}\}^{N}$
, Classifiers $C =\{c_{i}\}^{N}$, Multi-classifiers model $M$        \\
\hspace*{0.02in} {\bf Output:}
Predicted labels $Y$
\begin{algorithmic}[1]
\State $Y := $ None
\For{i from 1 to $N$}
\State logit = $M$.getSoftmaxOutputs($X$, i)
\If{max(logits) > $\sigma_{i}$}
\State $Y$ $:=$ argmax(logits)
\State Break
\EndIf
\EndFor
\If{$Y = $~None}
    \State $Y$ $:=$ $M$.getEnsemblePrediction()
\EndIf
\State \Return Prediction
\end{algorithmic}
\end{algorithm}
To address this problem and further enhance the performance of shallow classifiers, attention modules are utilized to obtain classifier-specific features from the sharing backbone neural network. Inspired by RAN~\cite{attentionresidual}, we propose a simplified attention modules including one convolution layer for downsampling and one deconvolution layer for upsampling. A sigmoid activation is attached after attention modules to obtain attention maps between 0 and 1. Then, the attention maps are involved in a dot product operation with origin features, yielding classifier-specific features. Its forward computation can be formulated as
\begin{equation}
Attention~Maps(W_{conv}, W_{deconv}, F) = \sigma(\phi(\psi(F, W_{conv})), W_{deconv}) \label{equation:attention}
\end{equation}
where $\psi$ and $\phi$ denote convolution function and deconvolution function respectively. $F$ represents the input features and $\sigma$ signifies a sigmoid function. Notes that batch normalization and ReLU activation function after convolution and deconvolution layers are omitted here.

Experiments results demonstrate that attention modules in SCAN lead to dramatic accuracy boost in shallow classifiers. For instance, 5.46\%, 4.13\% ,and 5.16\% accuracy gain can be observed on the shallow classifiers in ResNet50 on CIFAR100, compared with self distillation~\cite{selfdistillation}.
\subsection{Scalable inference mechanism\label{scalable inference mechanism}}
It is generally acknowledged that the prediction of neural networks with a higher confidence (softmax value) is more likely to be right. In this paper, we exploit this observation to determine whether a classifier gives a right or wrong prediction.
As described in Algorithm \ref{algorithm1}, we set different thresholds for shallow classifiers. If the maximal output of softmax in shallow classifier is larger than the corresponding threshold, its results will be adopted as the final prediction. Otherwise, the neural networks will employ a deeper classifier to predict, until the deepest one or the ensemble prediction. Because most of the computation for shallow classifiers is included by that for deep classifiers, there is no much extra computation introduced.

However, threshold controlled scalable inference causes another problem, that is the choice of thresholds for difference classifiers. Thresholds matters: (i) A lower threshold for shallow classifiers results in that most samples will be predicted by shallow classifiers, indicating more rapid response yet lower accuracy. (ii) Similarly, a higher threshold leads to a phenomenon that most samples will be determined by deeper classifiers, indicating precise prediction yet longer response time. (iii) By adjusting the value of thresholds, flexible accuracy-response time trade-offs can be approached on the fly. Instead of designing thresholds manually, we propose a genetic algorithm based method to search the most optimal thresholds as is depicted in Algorithm \ref{algorithm1}.

\textbf{Genes Coding:} Genes, a binary sequence is ought to be decoded into its corresponding threshold. To guarantee the accuracy of the model, we empirically restrict the lower bound for thresholds as 0.70. Its decode function can be formulated as
\begin{equation}
\sigma_{i} = 1 - \frac{0.3}{N} \cdot \sum_{n=1}^{N} S(n)=1\label{equation:decoding}
\end{equation}
Here $S(n)$ indicates the $n_{th}$ bit in the sequence of genes. $\sigma$ denotes the threshold of $i_{th}$ gene. $N$ is the length of binary sequence utilized to express one threshold. The more bits "1" there are in this sequence, the lower the threshold is.

\textbf{Fitness:} Another crucial issue is to choose the metrics for computing the fitness for thresholds. Targeting at accelerating the models and improving the performance at the same time, these two elements are taken into consideration for the fitness metrics, which can be formulated as
\begin{equation}
fitness = acceleration~ratio + \beta \cdot (accuracy-baseline)\label{equation:fitness}
\end{equation}
where $\beta$ is a hyper-parameter to balance the impact of these two elements. Adjustment of $\beta$ leads to trade-offs between accuracy and acceleration.

\begin{algorithm}[t]
\label{algotirhtm2}
\caption{Threshold Searching}
\hspace*{0.02in} {\bf Input:}  \hspace*{0.02in}
Genes $G$, Multi-classifiers model $M$, Dataset $D$ , Generations $g$\\
\hspace*{0.02in} {\bf Output:}
Optimal Thresholds $\Sigma = \{\sigma_{i}\}^{N}$
\begin{algorithmic}[1]
\State RandomlyInitialize($G$)
\For{i from 1 to $g$}
\State fitness $=$ getFitness($G$, $D$, $M$)~~~//~Calculate fitness of each gene according to Equation \ref{equation:fitness}.
\State $G:=$ weightedSelect($G$, fitness)~~~~~~//~Drop the genes with low fitness.
\State $G:=$ crossover($G$)~~~~~~~~~~~~~~~~~~~~~~~~~~~//~Each two genes cross over, generating new genes.
\State $G:=$ mutate($G$)~~~~~~~~~~~~~~~~~~~~~~~~~~~~~~~//~Each bit of genes may mutate with a low possibility.
\EndFor
\State $\Sigma :=$ decode($G$)~~~~~~~~~~~~~~~~~~~~~~~~~//~Decode the genes into thresholds according to Equation \ref{equation:decoding}.
\State \Return $\Sigma$
\end{algorithmic}
\end{algorithm}

\section{Experiments\label{section_experiment}}
We evaluate SCAN on two benchmark datasets: CIFAR100~\cite{cifar} and ImageNet (ILSVRC2012)~\cite{imagenet} and three kinds of neural networks with difference depth and width: VGG~\cite{vgg}, ResNet~\cite{resnet} and Wide ResNet~\cite{wideresnet}. During training periods, common techniques like data argumentation (random cropping and horizon flipping), learning rate decay, $l_2$ regularization are equipped. To fit the size of tiny images in CIFAR, we slightly adjust the kernel size and strides of convolution and pooling layers. The recommended value for hyper-parameters $\lambda$ and $\alpha$ in Equation \ref{equation1}, and $N$ in Equation \ref{equation:decoding} are 0.5, 5e-7 and 30, respectively. Note that the reported ImageNet (ILSVRC2012) accuracy is evaluated on validation set. All the experiments are conducted by PyTorch1.0 on GPU devices.

\begin{figure}
\begin{minipage}[t]{0.45\linewidth}
\centering
\includegraphics[height=5.1cm,width=6.8cm]{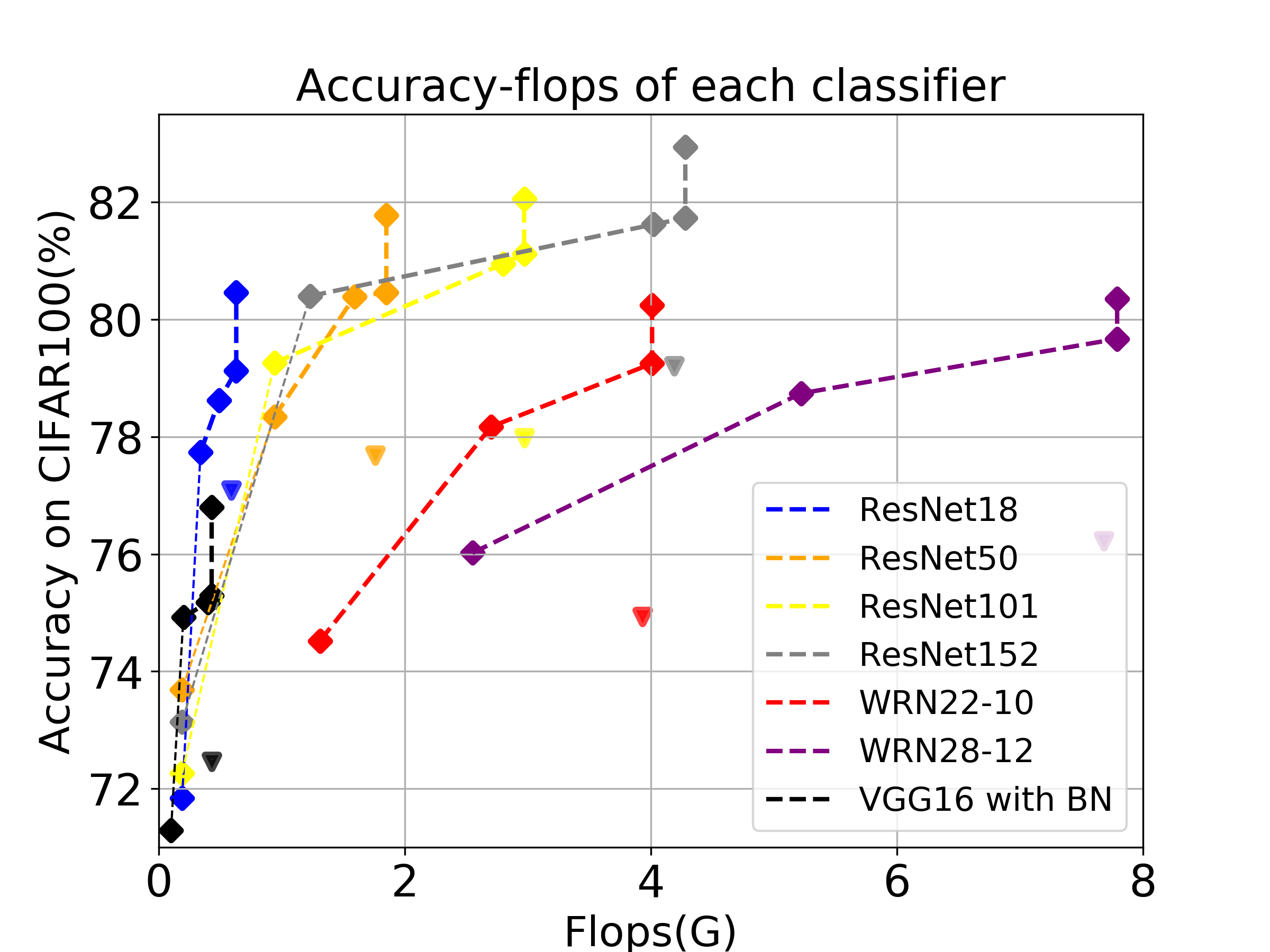}
\caption{Accuracy and computation of \newline each classifier on CIFAR100.}
\label{figure3}
\end{minipage}%
\begin{minipage}[t]{0.45\linewidth}
\centering
\includegraphics[height=5.1cm,width=6.8cm]{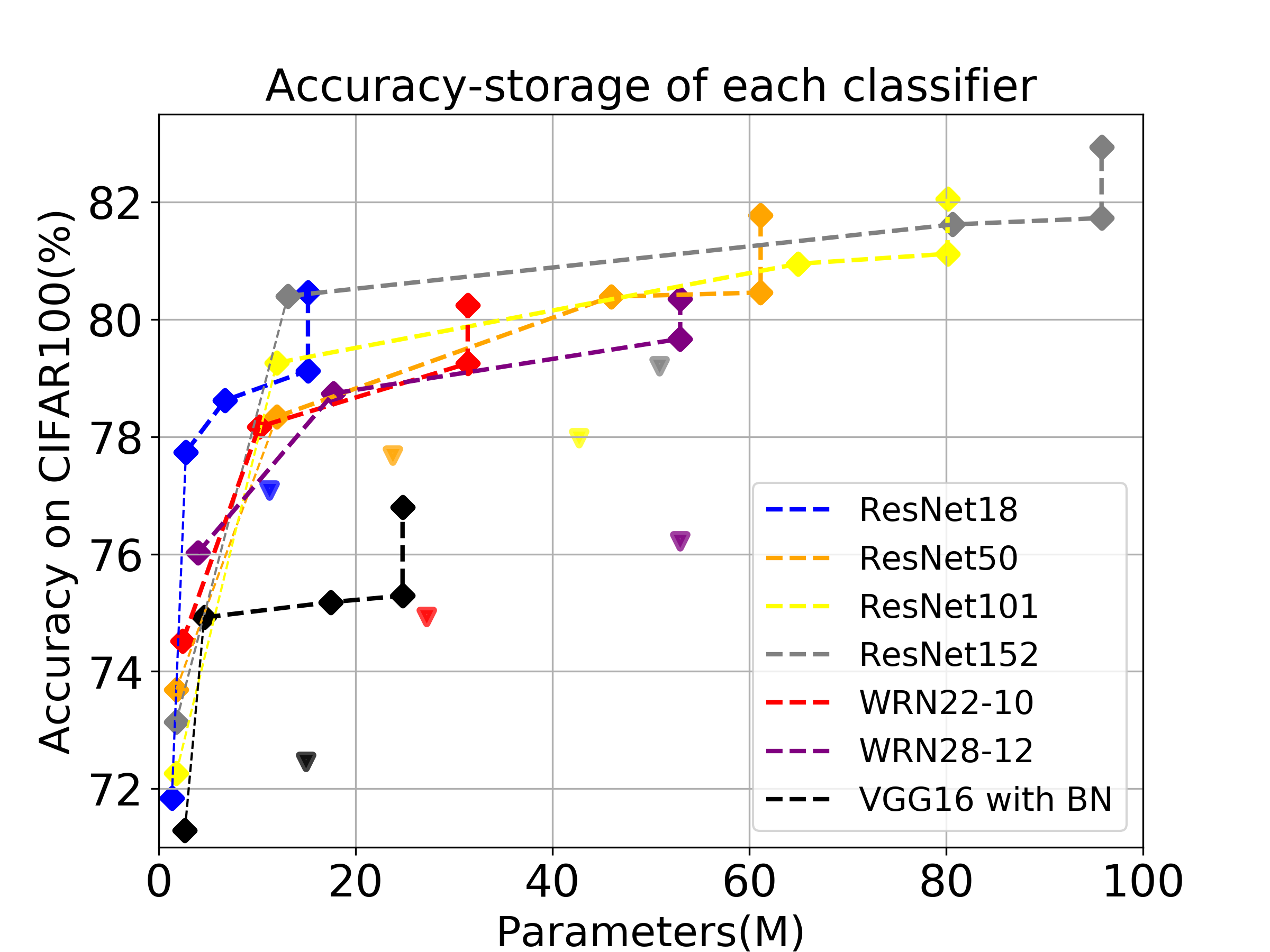}
\caption{Accuracy and storage of each classifier on CIFAR100.}
\label{figure4}
\end{minipage}
\end{figure}

\subsection{Results on CIFAR100}
Experiments results on CIFAR100 are depicted in Figure \ref{figure3} and \ref{figure4}. The squares connected in the same line from left to right denote the classifier\nicefrac{1}{4} to classifier\nicefrac{4}{4} and their ensemble. All of them share the same backbone neural network. The triangle with the same color out of lines denotes the corresponding baseline.

It is observed that (i) In all the situations, the classifier\nicefrac{2}{4} equipped with SCAN outperforms its baseline. (ii) 2.17X acceleration and 3.20X compression have been achieved on average with no accuracy drop. (iii) Compared with the corresponding baseline, 4.05\% accuracy increment can be obtained with 4.4\% relative computation increment on average. (iv) The ensemble of all classifiers in one neural network leads to 1.11\% performance gain with almost no incremental computation and storage. (v) Shallow classifiers benefit more from SCAN than deeper classifiers. (vi) Deeper or wider neural networks benefit more from SCAN than shallower or thinner neural networks.


\begin{table*}
\begin{center}
\caption{Experiments results of accuracy (\%) on CIFAR100.}
\begin{tabular}{|c|c|c|c|c|c|c|c|c|}
\hline
Models& Baseline & Classifier\nicefrac{1}{4}& Classifier\nicefrac{2}{4} &Classifier\nicefrac{3}{4}& Classifier\nicefrac{4}{4}& Ensemble\\
\hline
VGG16(BN)&72.46 & 71.29 & 74.92 & 75.18& 75.29& 76.80 \\
\hline
VGG19(BN)&72.25& 71.52& 74.02& 74.15& 74.43& 75.43 \\
\hline
ResNet18&77.09 &71.84 &77.74 &78.62&79.13&80.46\\
\hline
ResNet50&77.68 &73.69&78.34&80.39&80.45&81.78\\
\hline
ResNet101&77.98  &72.26&79.26&80.95&81.12&82.06\\
\hline
ResNet152&79.21  &73.14&80.40&81.73&81.62&82.94\\
\hline
WRN20-8&74.61&74.52 &78.17 &79.25&/&80.04\\
\hline
WRN44-8&76.22&76.02  &78.74&79.67&/&80.35\\
\hline
\end{tabular}
\end{center}
\label{table: accuracy on CIFAR100}
\end{table*}

\begin{table*}
\begin{center}
\caption{Experiments results of accuracy (\%) on ImageNet.}
\begin{tabular}{|c|c|c|c|c|c|c|c|c|}
\hline
Models& Baseline & Classifier\nicefrac{1}{4} &Classifier\nicefrac{2}{4} &Classifier\nicefrac{3}{4} &Classifier\nicefrac{4}{4}\\
\hline
ResNet18&68.02 &48.25 &58.00 &65.32&69.32\\
\hline
ResNet50&74.47& 53.86 & 66.54 &73.57 &75.88\\
\hline
ResNet101&75.24& 52.32 & 65.33 & 74.51&76.32\\
\hline
\end{tabular}
\end{center}
\label{table: accuracy on ImageNet}
\end{table*}




\begin{figure}
\vspace{-0.25cm}
\begin{minipage}[t]{0.45\linewidth}
\centering
\includegraphics[height=4.383cm,width=6.138cm]{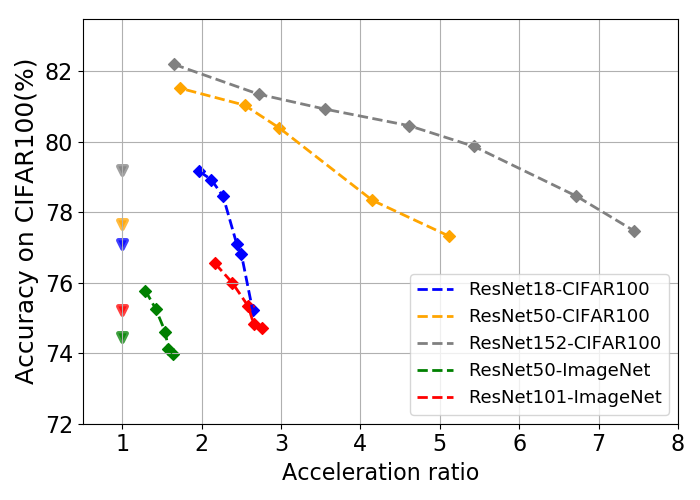}
\caption{Acceleration ratio and accuracy \newline of different thresholds.}
\label{fig:threshold}
\end{minipage}%
\begin{minipage}[t]{0.45\linewidth}
\centering
\includegraphics[height=4.4cm,width=5.4cm]{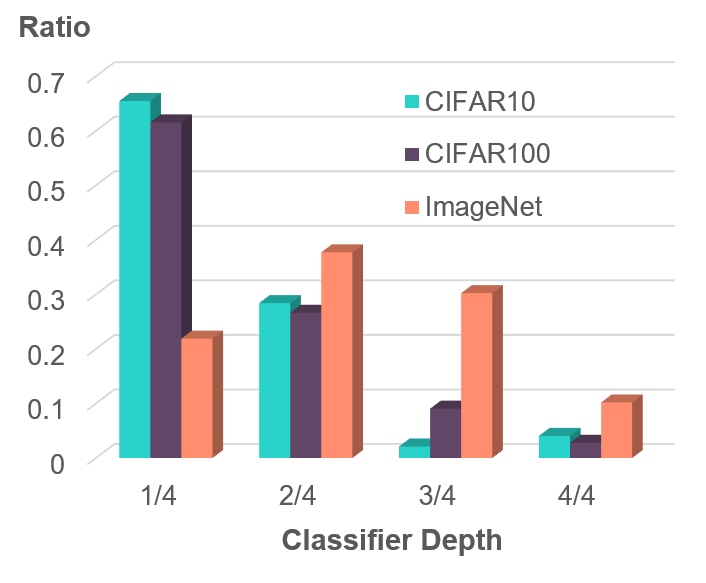}
\caption{Statistics of samples predicted in each classifier on three kinds of datasets.}
\label{ratio}
\end{minipage}
\end{figure}
\subsection{Results on ImageNet}
As shown in Table \ref{table: accuracy on ImageNet}, the same tendency of accuracy increment on all the classifiers can also be observed: (i) On average, 1.26\% accuracy gain on ImageNet can be achieved, varying from 1.41\% on ResNet50 as maximum to 1.08\% on ResNet101 as minimum. (ii) The depth of classifiers impacts their accuracy more significantly, which indicates there is less redundancy in ImageNet compared with CIFAR100.


\subsection{Results of scalable inference}
Experiments results for scalable inference are depicted in Figure \ref{fig:threshold}. The X axis is the acceleration ratio compared to its baseline. The Y axis is the top 1 accuracy evaluated on CIFAR100 and ImageNet. The squares connected in the same line denotes the results of different thresholds for the same neural network. The triangles on x=1 denote the baselines of different models.

It is observed that:(i) Compared with deploying one of the shallow classifier individually, scalable inference leads to a higher acceleration ratio. (ii) Compared with baselines on CIFAR100, 4.41X acceleration can be achieved with no drop on accuracy on average, varying from 2.41X on ResNet18 as minimum to 6.23X on ResNet152 as maximum. (iii) Compared with baselines on ImageNet, 1.99X acceleration can be achieved with no accuracy drop on average, varying from 1.54X on ResNet50 as minimum to 2.43X on ResNet101 as maximum.
(iv) More accuracy gain can be observed on deeper neural networks which is in accordance with the general observation that over-parameterized models have more potential to be compressed and accelerated.

\begin{figure}[htbp]
\vspace{-0.25cm}
    \centering
    \includegraphics[width=13.11cm,height=7.7cm]{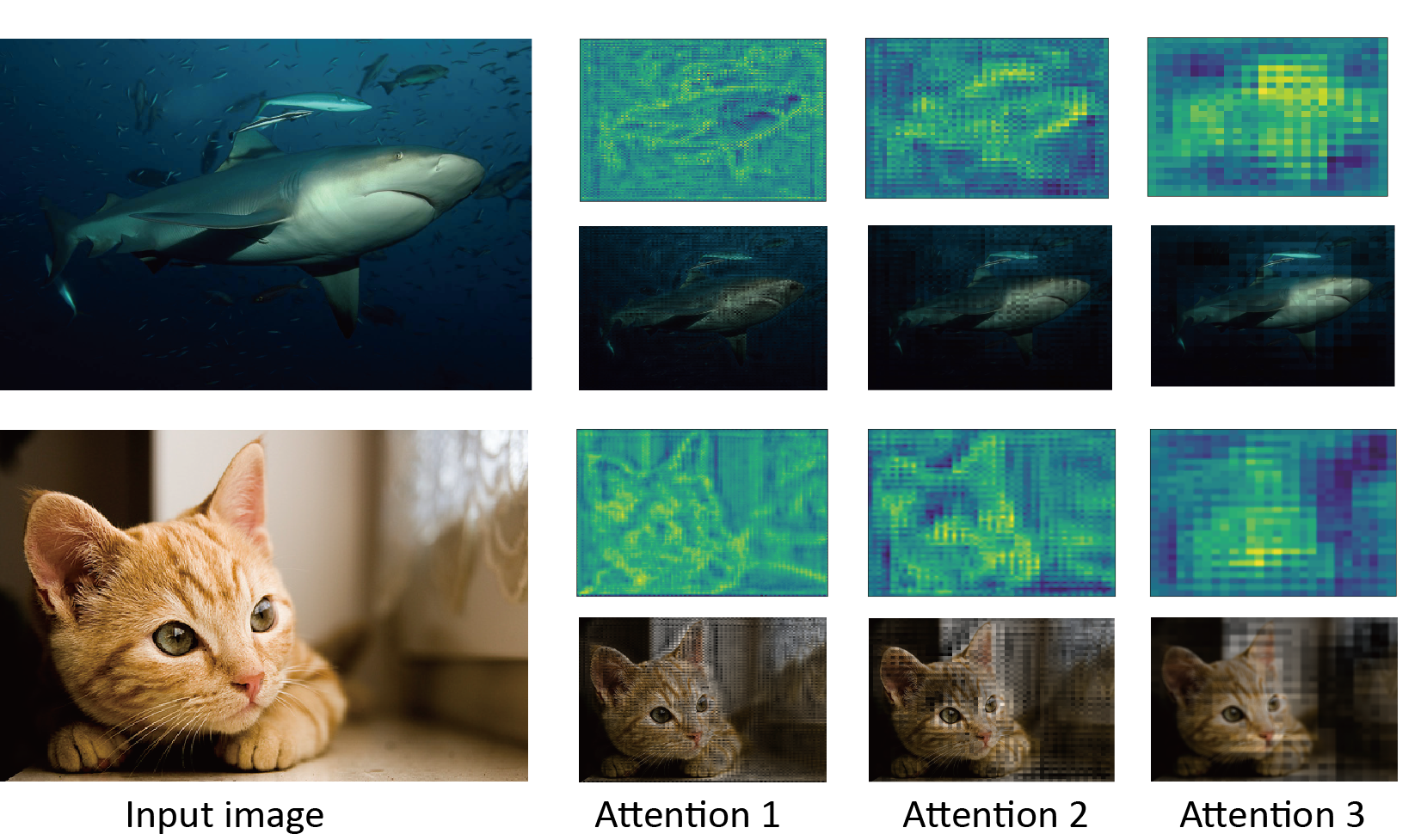}
    \caption{Visualization of attention maps in shallow classifiers.}
    \label{fig:attention}
\end{figure}

\section{Discussion\label{section_discusstion} }
\textbf{What have attention modules learned?}
In SCAN, attention modules are introduced to obtain classifier-specific features, leading to significant performance gain on shallow classifiers. We further visualize the spatial attention maps as depicted in Figure \ref{fig:attention}. The heat maps indicate learned attention maps, where the value of each pixel is computed as the mean value of pixels in the same position of all channels.

As is depicted in Figure \ref{fig:attention}, all the classifiers pay their attention on the same spatial position - the bodies of a shark and a cat, while ignoring the backgrounds, which indicates that all of the attention modules have learned to find the most informative pixels. The attention maps in classifiers \nicefrac{1}{4} seems to concentrate on the details of shark's and cat's features such as their outlines. In contrast, the attention maps in deeper classifiers \nicefrac{3}{4} focus more on the texture features, which indicates deep classifiers that have a larger receptive filed are more likely to predict based on global and low frequency information while shallow classifiers incline to be dominated by local and high frequency information.

\textbf{How many samples are predicted by shallow classifiers?}
The determinants of the acceleration ratio in SCAN is the number of samples predicted by shallow classifiers, which varies from thresholds, datasets and neural networks. Figure \ref{ratio} shows the statistics of samples predicted by each classifier of ResNet18 on CIFAR10, CIFAR100 and ImageNet with the same thresholds. It's observed that:(i) More than half samples in CIFAR10 and CIFAR100 can be classified in the shallowest classifier, which consumes the least computation compared with others. (ii) In ImageNet, more samples have to be predicted in the last two classifiers, which indicates the classification of ImageNet data is beyond the capacity of shallow classifiers. Based on these observation, two possible usage are proposed as follows:

Firstly, the number of samples predicted in different classifier can be utilized as a guidance of models compression. For example, the classifier\nicefrac{4}{4} in Figure \ref{ratio} provides extremely little valid prediction in CIFAR10 and CIFAR100, indicating there is much more redundancy and compression potential.

Secondly, it also can be utilized as a metric if datasets difficulty. It's not rigorous to measure the difficulty of datasets by prediction accuracy because different datasets consist of different numbers of categories. SCAN provides a possible solution - the ratio of samples predicted by shallow classifiers.

\textbf{Future works:} Although SCAN has achieved significant acceleration and boost on accuracy, we still believe it has more potential. Firstly, more creditable judgement of whether the prediction of shallow classifiers should be adopted remains to be explored. This issue can be formulated as a binary classification problem which may be addressed by machine learning algorithms.

Secondly, continued optimization on the structure of shallow classifiers is necessary. The success of attention modules in SCAN proves that tiny adjustment on shallow classifiers can lead to dramatic accuracy boost, indicating that more compact and efficient shallow classifiers can be achieved by a well designed structure.

\section{Conclusion\label{section_conclusion}}
We have proposed a novel neural networks training and inference framework named SCAN, whose benefits can be seen in three folds: Firstly, self distillation and attention modules are utilized to train compact and efficient shallow classifiers to achieve static acceleration and compression. Secondly, SCAN exploits the diversity in prediction difficulty to accomplish human-like sample-specific conditional execution, yielding scalability and a high acceleration ratio. Thirdly, compared with self distillation and its corresponding baseline, more significant accuracy gain can be achieved on the ensemble of all classifiers with a negligible growth on computation and storage.
{
\small
\bibliographystyle{plain}
\bibliography{ref}
}
\end{document}